\documentclass[11pt]{article}

\usepackage[final]{acl}

\definecolor{darkblue}{HTML}{000099}
\usepackage{fontspec}
\usepackage[english]{babel}
\usepackage{microtype}
\usepackage{amsmath,amssymb,amsthm}
\usepackage{booktabs,tabularx}
\usepackage{graphicx}
\usepackage{algorithm}
\usepackage{algpseudocode}
\usepackage{tikz}
\usetikzlibrary{arrows.meta,positioning,calc}
\usepackage{url}

\newcommand{\method}{CAST}
\newcommand{\R}{\mathbb{R}}
\newcommand{\norm}[1]{\left\lVert#1\right\rVert}

\newcommand{\PowerTop}{\operatorname{PowerTop}}

\title{CAST: History-Adaptive Spectral Clipping for LLM Pretraining}

\author{
    Jiacheng Li,
    \ Jianchao Tan\thanks{Corresponding author.},\\
    \ {\bf Kefeng Zhang,}
    \ {\bf Yuchen Xie,}
    \ {\bf Xunliang Cai}\\
    \ {\bf Meituan, Beijing, China} \\
    \texttt{\{lijiacheng14, tanjianchao02\}@meituan.com}
}

\begin{document}
\maketitle

\newcommand{\FullMainRows}{%
\textbf{CAST} & \textbf{1.6516} & \textbf{1.8738} & --- \\
GlobalGC & 1.6718 & 1.8962 & 19/20 \\
AdaGC & 1.6625 & 1.8872 & 20/20 \\
SSC-constant & 1.6638 & 1.8864 & 20/20 \\
SSC-cosine & 1.6700 & 1.8947 & 19/20 \\
GradientStabilizer & 1.6828 & 1.9034 & 18/20 \\
GradPower & 1.6823 & 1.9139 & 20/20 \\
ClipByValue & 1.6789 & 1.9022 & 20/20 \\
AGC & 1.6641 & 1.8856 & 19/20 \\
ZClip & 1.6711 & 1.8937 & 19/20 \\
}

\newcommand{\FullAblationRows}{%
CAST & \textbf{1.6516} & \textbf{1.8738} & 0.884 \\
w/o HS & 1.6541 & 1.8747 & 0.403 \\
\shortstack[l]{w/\ $\tau=0.1$} & 1.6937 & 1.9215 & 1.000 \\
\shortstack[l]{w/\ cosine $\tau$} & 1.6767 & 1.9034 & 1.000 \\
}

\newcommand{\FullEfficiencyRows}{%
Mean (s) & 44.408 & 43.533 & 43.580 \\
Median (s) & 44.206 & 43.329 & 43.375 \\
P95 (s) & 44.876 & 43.960 & 44.119 \\
\midrule
\begin{tabular}[c]{@{}l@{}}Tokens/s/\\device\end{tabular} & 5903 & 6022 & 6015 \\
}

\newcommand{\FullStabilityRows}{%
CAST & 0.1510 & 0.3702 & 0.5476 & \textbf{0} & \textbf{1} \\
GlobalGC & 0.1541 & 0.4574 & 15.1610 & 10 & 38 \\
AdaGC & 0.1343 & 0.3485 & 1.0541 & \underline{1} & 6 \\
SSC-constant & 0.1591 & 0.4723 & 1.1829 & \underline{1} & 22 \\
SSC-cosine & 0.1653 & 0.4996 & 1.1922 & 2 & 21 \\
GradientStabilizer & 0.1701 & 0.6413 & 4.1736 & 5 & 31 \\
GradPower & 0.1142 & 0.6987 & 4.9144 & 25 & 422 \\
ClipByValue & 0.1645 & 0.6814 & 5.0252 & 22 & 58 \\
AGC & 0.1425 & 0.4701 & 3.9416 & 25 & 52 \\
ZClip & 0.1540 & 1.2761 & 918.7862 & 52 & 75 \\
\midrule
CAST w/o HS & 0.1463 & 0.3382 & 0.4548 & \textbf{0} & 3 \\
CAST w/ $\tau=0.1$ & 0.1207 & 2.7853 & 4.0557 & 194 & 6 \\
CAST w/ cosine $\tau$ & 0.1083 & 0.8505 & 1.6856 & 21 & \underline{2} \\
}

\newcommand{\FullDomainRows}{%
01 Academic / vocational & \textbf{2.2466} & 2.2718 & \underline{2.2569} & 2.2601 & 2.2693 & 2.2849 & 2.2943 & 2.2801 & 2.2603 & 2.2716 \\
02 Clinical / health & \textbf{1.9951} & 2.0179 & \underline{2.0054} & 2.0078 & 2.0168 & 2.0304 & 2.0401 & 2.0255 & 2.0082 & 2.0181 \\
03 Consumer / leisure & \textbf{2.1824} & 2.2031 & \underline{2.1924} & 2.1935 & 2.2026 & 2.2195 & 2.2273 & 2.2124 & 2.1938 & 2.2050 \\
04 Economy / services & \textbf{2.0175} & 2.0434 & 2.0310 & \underline{2.0309} & 2.0399 & 2.0575 & 2.0645 & 2.0512 & 2.0329 & 2.0422 \\
05 IT / programming & \textbf{1.8519} & 1.8727 & \underline{1.8631} & 1.8646 & 1.8713 & 1.8864 & 1.8941 & 1.8807 & 1.8634 & 1.8758 \\
06 Periodicals & \textbf{2.7619} & 2.7855 & 2.7760 & \underline{2.7744} & 2.7822 & 2.7960 & 2.8002 & 2.7914 & 2.7766 & 2.7865 \\
07 Regional culture & \textbf{2.1047} & 2.1263 & \underline{2.1155} & 2.1164 & 2.1243 & 2.1391 & 2.1479 & 2.1330 & 2.1168 & 2.1267 \\
08 Science / technology & \textbf{1.8652} & 1.8862 & \underline{1.8752} & 1.8782 & 1.8852 & 1.8992 & 1.9072 & 1.8944 & 1.8773 & 1.8881 \\
09 Sports / vehicles & \textbf{2.2699} & 2.2944 & 2.2844 & \underline{2.2838} & 2.2921 & 2.3096 & 2.3162 & 2.3025 & 2.2840 & 2.2947 \\
10 CMMLU text & \underline{1.4037} & 1.4355 & 1.4218 & 1.4363 & 1.4444 & \textbf{1.4010} & 1.4414 & 1.4344 & 1.4142 & 1.4178 \\
11 Open-source code A & \textbf{0.9368} & 0.9520 & 0.9450 & 0.9433 & 0.9495 & 0.9586 & 0.9604 & 0.9525 & \underline{0.9427} & 0.9534 \\
12 Open-source code B & \textbf{0.9367} & 0.9518 & 0.9449 & 0.9432 & 0.9493 & 0.9585 & 0.9603 & 0.9524 & \underline{0.9426} & 0.9532 \\
13 English books & \textbf{2.9816} & 3.0043 & 2.9952 & \underline{2.9944} & 2.9987 & 3.0211 & 3.0242 & 3.0098 & 2.9957 & 3.0059 \\
14 FullStackBench text & \textbf{1.1398} & 1.1726 & 1.1725 & \underline{1.1415} & 1.1682 & 1.1778 & 1.1702 & 1.1694 & 1.1526 & 1.1626 \\
15 GSM8K text & \textbf{0.7507} & 0.7656 & 0.7526 & 0.7604 & 0.7587 & 0.7760 & 0.7770 & 0.7680 & \underline{0.7510} & 0.7664 \\
16 HumanEval text & 1.1460 & 1.1416 & 1.1612 & 1.1466 & \underline{1.1357} & 1.1651 & 1.1835 & 1.1718 & \textbf{1.1351} & 1.1394 \\
17 MMLU text & \underline{1.4045} & 1.4435 & 1.4355 & 1.4263 & 1.4555 & \textbf{1.3934} & 1.4597 & 1.4359 & 1.4423 & 1.4108 \\
18 English web & \textbf{2.7558} & 2.7810 & \underline{2.7697} & 2.7709 & 2.7797 & 2.7981 & 2.8058 & 2.7888 & 2.7716 & 2.7816 \\
19 Chinese essays & \textbf{3.9568} & 3.9867 & \underline{3.9759} & 3.9789 & 3.9880 & 4.0075 & 4.0102 & 3.9984 & 3.9766 & 3.9908 \\
20 Software engineering & \textbf{0.7679} & 0.7873 & \underline{0.7703} & 0.7767 & 0.7831 & 0.7882 & 0.7939 & 0.7918 & 0.7737 & 0.7830 \\
}

\newcommand{\FullDomainRowsA}{%
01 Academic / vocational & \textbf{2.2466} & 2.2718 & \underline{2.2569} & 2.2601 & 2.2693 \\
02 Clinical / health & \textbf{1.9951} & 2.0179 & \underline{2.0054} & 2.0078 & 2.0168 \\
03 Consumer / leisure & \textbf{2.1824} & 2.2031 & \underline{2.1924} & 2.1935 & 2.2026 \\
04 Economy / services & \textbf{2.0175} & 2.0434 & 2.0310 & \underline{2.0309} & 2.0399 \\
05 IT / programming & \textbf{1.8519} & 1.8727 & \underline{1.8631} & 1.8646 & 1.8713 \\
06 Periodicals & \textbf{2.7619} & 2.7855 & 2.7760 & \underline{2.7744} & 2.7822 \\
07 Regional culture & \textbf{2.1047} & 2.1263 & \underline{2.1155} & 2.1164 & 2.1243 \\
08 Science / technology & \textbf{1.8652} & 1.8862 & \underline{1.8752} & 1.8782 & 1.8852 \\
09 Sports / vehicles & \textbf{2.2699} & 2.2944 & 2.2844 & \underline{2.2838} & 2.2921 \\
10 CMMLU text & \underline{1.4037} & 1.4355 & 1.4218 & 1.4363 & 1.4444 \\
11 Open-source code A & \textbf{0.9368} & 0.9520 & 0.9450 & 0.9433 & 0.9495 \\
12 Open-source code B & \textbf{0.9367} & 0.9518 & 0.9449 & 0.9432 & 0.9493 \\
13 English books & \textbf{2.9816} & 3.0043 & 2.9952 & \underline{2.9944} & 2.9987 \\
14 FullStackBench text & \textbf{1.1398} & 1.1726 & 1.1725 & \underline{1.1415} & 1.1682 \\
15 GSM8K text & \textbf{0.7507} & 0.7656 & 0.7526 & 0.7604 & 0.7587 \\
16 HumanEval text & 1.1460 & 1.1416 & 1.1612 & 1.1466 & \underline{1.1357} \\
17 MMLU text & \underline{1.4045} & 1.4435 & 1.4355 & 1.4263 & 1.4555 \\
18 English web & \textbf{2.7558} & 2.7810 & \underline{2.7697} & 2.7709 & 2.7797 \\
19 Chinese essays & \textbf{3.9568} & 3.9867 & \underline{3.9759} & 3.9789 & 3.9880 \\
20 Software engineering & \textbf{0.7679} & 0.7873 & \underline{0.7703} & 0.7767 & 0.7831 \\
}

\newcommand{\FullDomainRowsB}{%
01 Academic / vocational & 2.2849 & 2.2943 & 2.2801 & 2.2603 & 2.2716 \\
02 Clinical / health & 2.0304 & 2.0401 & 2.0255 & 2.0082 & 2.0181 \\
03 Consumer / leisure & 2.2195 & 2.2273 & 2.2124 & 2.1938 & 2.2050 \\
04 Economy / services & 2.0575 & 2.0645 & 2.0512 & 2.0329 & 2.0422 \\
05 IT / programming & 1.8864 & 1.8941 & 1.8807 & 1.8634 & 1.8758 \\
06 Periodicals & 2.7960 & 2.8002 & 2.7914 & 2.7766 & 2.7865 \\
07 Regional culture & 2.1391 & 2.1479 & 2.1330 & 2.1168 & 2.1267 \\
08 Science / technology & 1.8992 & 1.9072 & 1.8944 & 1.8773 & 1.8881 \\
09 Sports / vehicles & 2.3096 & 2.3162 & 2.3025 & 2.2840 & 2.2947 \\
10 CMMLU text & \textbf{1.4010} & 1.4414 & 1.4344 & 1.4142 & 1.4178 \\
11 Open-source code A & 0.9586 & 0.9604 & 0.9525 & \underline{0.9427} & 0.9534 \\
12 Open-source code B & 0.9585 & 0.9603 & 0.9524 & \underline{0.9426} & 0.9532 \\
13 English books & 3.0211 & 3.0242 & 3.0098 & 2.9957 & 3.0059 \\
14 FullStackBench text & 1.1778 & 1.1702 & 1.1694 & 1.1526 & 1.1626 \\
15 GSM8K text & 0.7760 & 0.7770 & 0.7680 & \underline{0.7510} & 0.7664 \\
16 HumanEval text & 1.1651 & 1.1835 & 1.1718 & \textbf{1.1351} & 1.1394 \\
17 MMLU text & \textbf{1.3934} & 1.4597 & 1.4359 & 1.4423 & 1.4108 \\
18 English web & 2.7981 & 2.8058 & 2.7888 & 2.7716 & 2.7816 \\
19 Chinese essays & 4.0075 & 4.0102 & 3.9984 & 3.9766 & 3.9908 \\
20 Software engineering & 0.7882 & 0.7939 & 0.7918 & 0.7737 & 0.7830 \\
}

\begin{abstract}
Soft spectral clipping (SSC) reduces gradient singular values, but its effect depends on the gradient scale and clipping threshold.
We introduce \textbf{CAST} (Clipping with Adaptive Scales and Thresholds), a history-adaptive extension of SSC.
AdamW's previous second moment sets row and column scales.
An exponential moving average of the clipped output's estimated spectral norm sets each matrix's threshold.
CAST restores the original coordinates before the AdamW update.
We train a mixture-of-experts (MoE) model with \textbf{19.08B} total parameters for \textbf{83.89B} tokens.
CAST achieves lower training and mean validation losses than nine gradient-processing baselines, with the fewest relative gradient spikes.
It reaches the best baseline's final smoothed training loss in \textbf{7.2\%} fewer steps.
\end{abstract}

\section{Introduction}

Gradient clipping\citep{zhang2019gradient} controls large gradients during language model\citep{meituanlongcatteam2025longcatflashtechnicalreport,brown2020language,liu2024deepseek,yang2025qwen3,deepseekai2026deepseekv4,kimiteam2025kimik2openagentic,kimiteam2026kimik3openfrontier} pretraining \citep{pascanu2013}.
When applied before AdamW, it also changes the moment estimates used by later updates \citep{adamw2019}.
The clipping rule therefore affects both the current update and the optimizer's state over time.

\begin{figure}[t]
\centering
\includegraphics[width=0.9\columnwidth]{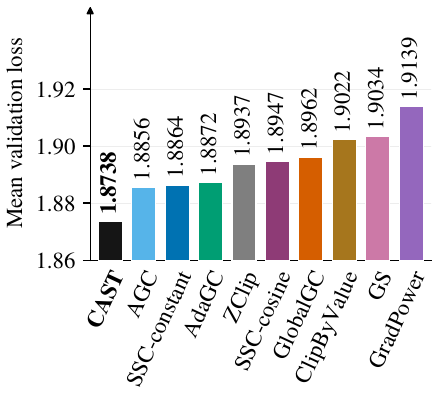}
\caption{Validation loss after 5,000 steps: unweighted mean over 20 subsets (one run per method; seed 48). GS: GradientStabilizer. The vertical axis is truncated. Full comparison: Table~\ref{tab:main}.}
\label{fig:headline}
\end{figure}

Global norm clipping rescales the whole gradient by a single factor.
This limits the influence of an unusually large gradient while preserving its direction.
A matrix gradient can be large mainly along a few directions.
Rescaling the entire matrix then also weakens other directions, even when their magnitudes are already moderate.
Spectral clipping controls the gradient magnitude in each singular direction.
SPECTRA\citep{spectra2026} develops a smooth spectral transformation for application before or after a base optimizer\citep{adamw2019,kingma2014adam,jordan2024muon}.
We use pre-optimizer SSC (SSC-Pre) because CAST also acts on gradients before AdamW.
We label its constant and cosine threshold schedules SSC-constant and SSC-cosine.

A clipping threshold is meaningful only relative to the gradient scale.
Persistent differences in row and column magnitudes affect the spectrum on which clipping acts.
An unusually large value in one part of a matrix may be routine in another, and the overall scale can change over training.
A fixed threshold may therefore apply different clipping strengths as the gradient scale changes.
SPECTRA's decaying pre-clipping threshold shows the value of adjusting this threshold during training.
Such a schedule depends on the training step.
A history-based rule can instead respond to changes in each matrix.
We therefore use each matrix's history to set its gradient scale and clipping threshold.

\begin{figure*}[t]
\centering
\begin{tikzpicture}[
 box/.style={draw=black!65,rounded corners=2pt,align=center,
   minimum height=10mm,font=\normalsize,inner xsep=0pt,inner ysep=2pt},
 arr/.style={-{Stealth[length=1.5mm]},semithick,draw=black!70},
 feedback/.style={arr,draw=blue!45!black,dashed,rounded corners=2pt},
 formula/.style={font=\normalsize,inner sep=0pt,anchor=base west}]
 \path[use as bounding box] (0,-1.38) rectangle (16,1.942);

 \node[box,fill=blue!5,minimum width=33mm] (scale) at (1.7,0) {};
 \node[font=\normalsize,inner sep=0pt] at (1.7,.24) {1. HS};
 \node[formula] (scale-prefix) at (.36,-.27) {$Z=G\oslash S($};
 \node[formula] (oldv) at (scale-prefix.base east) {$V_{\rm old}$};
 \node[formula] at (oldv.base east) {$)$};
 \node[box,fill=black!4,minimum width=32mm] (ssc) at (5.95,0)
   {SSC\\$Z\mapsto Z_{\rm out}$};
 \node[box,minimum width=34mm] (restore) at (10.25,0)
   {Restore scale\\$G_{\rm out}=S\odot Z_{\rm out}$};
 \node[box,minimum width=27mm] (adam) at (14.4,0) {Optimizer};
 \draw[arr] (scale.east)--(ssc.west);
 \draw[arr] (ssc.east)--(restore.west);
 \draw[arr] (restore.east)--(adam.west);

 \node[box,fill=blue!5,minimum width=34mm] (hist) at (5.95,1.42)
   {2. Output feedback\\$\tau=1.3h_{\rm old}$};
 \draw[arr] (hist.south)--(ssc.north);
 \draw[feedback] (8.02,0) |- (hist.east);
 \fill[black!70] (8.02,0) circle (.55pt);

 \node[box,fill=blue!5,minimum width=9mm,minimum height=5.5mm,
   inner ysep=0pt] (newv) at (14.4,-1.08) {$V$};
 \draw[arr] (adam.south)--(newv.north);
 \draw[feedback,shorten >=2pt] (newv.west) -| (oldv.south |- scale.south);
 \node[font=\normalsize,fill=white,inner xsep=4pt,inner ysep=1pt]
   at (8.02,-1.08) {next step};
\end{tikzpicture}
\caption{\method{} adds historical scaling (HS) and output feedback around SSC (Algorithm~\ref{alg:ssc}). AdamW uses restored $G_{\rm out}$; its $V$ becomes the next $V_{\rm old}$, bias-corrected within $S$. The EMA $h_{\rm old}$ tracks the estimated $\|Z_{\rm out}\|_2$. Dashed histories advance only after a successful update.}
\label{fig:overview}
\end{figure*}
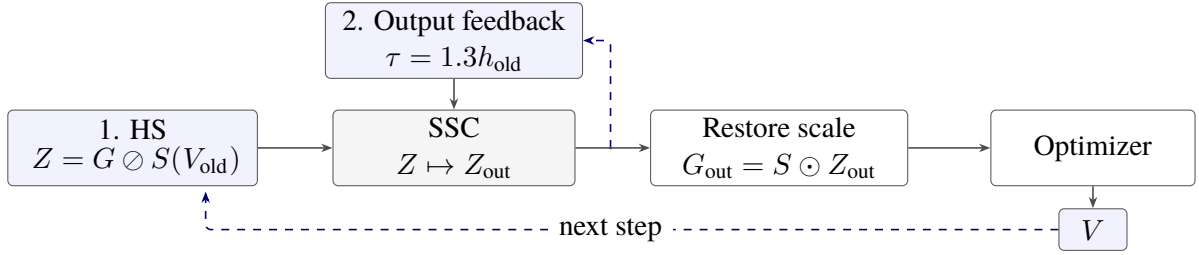

We introduce \method{} (Clipping with Adaptive Scales and Thresholds), which uses two complementary histories (Figure~\ref{fig:overview}).
Historical scaling (HS) uses row and column summaries of AdamW's previous second moment to set the scale for each gradient matrix.
SSC acts in these standardized coordinates, and the output returns to its original scale before updating AdamW.
AdamW updates both moments from this restored gradient.
The updated second moment supplies the scale used at the next step.

We also maintain an exponential moving average (EMA) of the clipped output's estimated spectral norm, computed in the same standardized coordinates.
This estimate measures the largest singular value after clipping.
The previous EMA sets the current threshold before clipping begins.
CAST retains SSC's soft map and Newton--Schulz routine.
The two histories control the input coordinates and clipping threshold.

Our contributions are threefold:
\begin{itemize}
\item \textbf{Historical scaling.} We use row and column summaries of the previous second moment to define the coordinates of spectral clipping.
We then restore the gradient's original scale.
\item \textbf{Output feedback.} We adapt each matrix's threshold to its clipped output history, so the threshold follows changes in the clipped spectrum.
\item \textbf{Pretraining results.} \method{} achieves the lowest training and mean validation losses among nine baselines in our MoE study (Figure~\ref{fig:headline}).
We examine its component effects, gradient dynamics, and training cost.
\end{itemize}

\section{Related Work}

\subsection{Spectral Clipping}

SPECTRA is the direct starting point for this work \citep{spectra2026}.
It smoothly reduces singular values through a matrix inverse square root, computed with Newton--Schulz iterations\citep{liu2025muon}.
Its framework combines spectral clipping of optimizer updates with optional gradient preprocessing.
This gradient preprocessing forms the SSC baselines in our study.

SPECTRA also studies how the clipping threshold changes during training.
Its pre-clipping study compares a fixed threshold of 0.1 with cosine decay from 10 to 0.01.
The decaying schedule gives better results.
We use the previous second moment to set the clipping scale and the complete clipped output to update the threshold.
We retain full-matrix soft clipping throughout.
Section~\ref{sec:setup} describes our SSC baselines.

\subsection{Gradient Transformations}

Adafactor represents second moments through row and column statistics \citep{adafactor2018}.
Our scale uses the same factored structure, with positive floors, to define the coordinates of spectral clipping.
AdamW keeps its full moment states and uses the gradient after we restore its original scale.

AdaGC adapts per-tensor thresholds from previous clipped norms \citep{adagc2025}.
Our spectral history follows this output-based principle for each matrix.
It uses an estimated spectral norm in standardized coordinates.
Parameters outside the spectral branch use an output-norm controller.
ZClip uses running global norm statistics to detect spikes \citep{zclip2025}.
GradientStabilizer transforms global magnitude using two scalar moving averages \citep{gradientstabilizer2025}.
These histories track different quantities from AdamW's coordinatewise second moment.

Other transformations use current parameter or gradient magnitudes.
AGC bounds unitwise gradient norms relative to parameter norms \citep{agc2021}; coordinate clipping bounds individual entries.
GradPower applies a signed elementwise power before the optimizer \citep{gradpower2025}.
These methods cover global, tensor, coordinate, and spectral gradient processing.

\section{Soft Spectral Clipping}
\label{sec:background}

For a singular value $\sigma\geq0$ and threshold $\tau>0$, the SPECTRA soft map is
\begin{equation}
 \phi_\tau(\sigma)=\frac{\sigma}{\sqrt{1+(\sigma/\tau)^2}}.
 \label{eq:phi}
\end{equation}
It slightly reduces small singular values and applies a stronger reduction to larger ones.
For $B\in\mathbb{R}^{q\times d}$ with $q\leq d$, its matrix form is
\begin{equation}
 H_\tau(B)
 =\tau(\tau^2I+BB^\top)^{-1/2}B.
 \label{eq:ssc}
\end{equation}
The inverse square root acts on the full $q\times q$ Gram matrix.
Newton--Schulz iterations approximate this matrix function without computing an explicit singular value decomposition \citep{spectra2026}.

Algorithm~\ref{alg:ssc} restates the SSC routine from SPECTRA in our notation.
It first bounds the largest squared singular value using the full Gram matrix.
If that bound is at or below the squared threshold, it returns the input unchanged.
Otherwise, it approximates $H_\tau$ with $K$ Newton--Schulz iterations.
We also return a Boolean $a$ that records whether clipping was applied.
This flag lets the caller preserve the original gradient exactly when SSC is bypassed.

\begin{algorithm}[t]
\caption{Soft spectral clipping (SSC) \citep{spectra2026}}
\label{alg:ssc}
\begin{algorithmic}[1]
\Require $B\in\R^{q\times d}$, $q\leq d$; threshold $\tau>0$; iterations $K$
\Ensure Output $B_{\rm out}$ and clipping flag $a$
\State $C\gets BB^\top$
\State $u\gets\min\{\|C\|_{\rm F},\max_i\sum_j|C_{ij}|\}$
\If{$u\leq\tau^2$}
 \State \Return $(B,\mathrm{false})$
\EndIf
\State $\alpha\gets\tau^2+u$; $A_0\gets(\tau^2I+C)/\alpha$
\State $Y\gets A_0$; $R\gets I$
\For{$j=1,\ldots,K$}
 \State $T\gets(3I-RY)/2$
 \State $Y\gets YT$; $R\gets TR$
\EndFor
\State \Return $((\tau/\sqrt\alpha)RB,\mathrm{true})$
\end{algorithmic}
\end{algorithm}

\begin{figure*}[t]
\centering
\includegraphics[width=\textwidth]{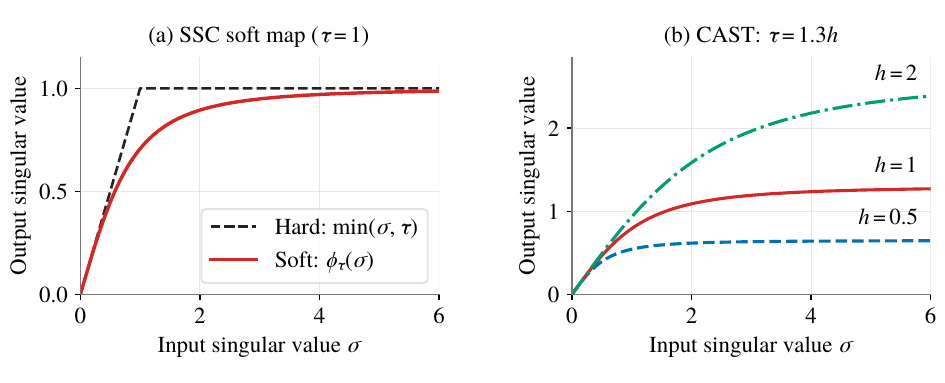}
\caption{The soft map and CAST's threshold choice. (a) Hard and soft clipping at $\tau=1$, following the functional view of SSC in \citet{spectra2026}. (b) The same soft map at $\tau=1.3h$ for three illustrative values of the previous output history $h$. These analytical curves describe the active branch in standardized coordinates; the bypass gate and finite Newton--Schulz approximation are separate steps in Algorithm~\ref{alg:ssc}.}
\label{fig:response}
\end{figure*}

The soft map in Equation~\ref{eq:ssc} describes the active clipping branch.
Figure~\ref{fig:response} illustrates its shape and the effect of changing the threshold.
CAST and both SSC baselines share the gate and the Newton--Schulz approximation.
Appendix~\ref{app:numerics} gives the input rescaling used to evaluate Algorithm~\ref{alg:ssc} safely in FP32.

\section{Method: \method{}}
\label{sec:method}

Let $G\in\mathbb{R}^{m\times n}$ be a gradient after accumulation, distributed reduction, and mixed-precision unscaling.
We transform it into $G_{\rm out}$ before either AdamW moment is updated:
\begin{align}
 M_t&=\beta_1M_{t-1}+(1-\beta_1)G_{\rm out},\nonumber\\
 V_t&=\beta_2V_{t-1}+(1-\beta_2)G_{\rm out}^{\odot2}.
 \label{eq:moments}
\end{align}
AdamW then applies its usual bias correction and decoupled weight decay.
We write $\odot$ and $\oslash$ for elementwise multiplication and division, and $\|\cdot\|_{\rm F}$ for the Frobenius norm.

Our method makes two changes around Algorithm~\ref{alg:ssc}: it standardizes the input and derives the threshold from output history.
SSC acts on the full matrix of each selected parameter.
We select two-dimensional tensors in the dense optimizer group, excluding embeddings and output matrices by name.
Other tensors, including parameters assigned to the expert optimizer, use the output-norm controller described below.
Appendix~\ref{app:scope} gives the selection rule for MoE projections and routers.

Algorithm~\ref{alg:method} separates these additions from the inherited SSC routine.
It shows matrices that have completed calibration and have a gradient at every step.
Appendix~\ref{app:state} covers initialization and missing gradients.
Both histories are read before the current AdamW update and committed only after it succeeds.

\begin{algorithm}[!t]
\caption{\method{} gradient preprocessing for AdamW}
\label{alg:method}
\begin{algorithmic}[1]
\Require Gradients $G$, old corrected moments $\widehat V$, histories $h$
\For{each selected matrix after calibration}
 \Statex \hspace{\algorithmicindent}\textit{1. HS}
 \State $S\gets\operatorname{Scale}(\widehat V)$ \Comment{Equation~\ref{eq:scale}}
 \State $Z\gets G\oslash S$; $B\gets\operatorname{Orient}(Z)$
 \Statex \hspace{\algorithmicindent}\textit{2. Threshold from output history}
 \State $\tau\gets1.3h$
 \State $(B_{\rm out},a)\gets\operatorname{SSC}(B,\tau,5)$ \Comment{Algorithm~\ref{alg:ssc}}
 \State $Z_{\rm out}\gets\operatorname{Unorient}(B_{\rm out})$
 \State $G_{\rm out}\gets G$
 \If{$a=\mathrm{true}$}
  \State $G_{\rm out}\gets S\odot Z_{\rm out}$
 \EndIf
 \State $s\gets\PowerTop(Z_{\rm out},3)$
 \If{$s>0$}
  \State Stage $h'\gets0.95h+0.05s$
 \EndIf
\EndFor
\State Clip other tensors with the historical norm rule
\State Apply AdamW to all transformed gradients
\State Commit staged state after a successful update
\end{algorithmic}
\end{algorithm}

\subsection{Historical Scaling}
\label{sec:scale}

We use the previous second moment to measure each gradient relative to its historical coordinate scales.
After $k\geq1$ moment updates, let $\widehat V=V/(1-\beta_2^k)$ denote the bias-corrected previous second moment.
Its row, column, and overall means are
\begin{equation}
 \begin{aligned}
 r_i&=\frac1n\sum_j\widehat V_{ij},\quad
 c_j=\frac1m\sum_i\widehat V_{ij},\\
 b_0&=\frac1{mn}\sum_{i,j}\widehat V_{ij}.
 \end{aligned}
\end{equation}
With $\rho=0.01$ and $\epsilon_s=10^{-8}$, define
\begin{align}
 b&=\max\{b_0,(\epsilon_s/\rho)^2\},\nonumber\\
 S_{ij}&=\sqrt{\frac{\max(r_i,\rho b)\max(c_j,\rho b)}{b}},
 \label{eq:scale}\\
 Z&=G\oslash S.\nonumber
\end{align}
The floors provide a positive reference for sparsely active coordinates.
The row and column factors account for persistent differences in gradient magnitudes.
Each coordinate uses one row factor and one column factor.
After clipping, we restore the original coordinates with $G_{\rm out}=S\odot Z_{\rm out}$.

\subsection{Adaptive Thresholds}
\label{sec:history}

We adapt the clipping threshold separately for each selected gradient matrix.
Each matrix maintains a positive scalar history $h$ of its clipped output's estimated spectral norm.
The current threshold uses its previous value:
\begin{equation}
 \tau_t=\lambda h_{t-1},\qquad\lambda=1.3.
 \label{eq:threshold}
\end{equation}
We orient $Z$ so that $B$ has the smaller dimension as its row count, then call Algorithm~\ref{alg:ssc} with $K=5$.
When the algorithm returns $a=\mathrm{false}$, we keep the original $G$ directly.
Otherwise, we restore the clipped matrix to $Z$'s orientation and set $G_{\rm out}=S\odot Z_{\rm out}$.

The controller observes the complete standardized output through three power iterations:
\begin{align}
 s_t&=\PowerTop(Z_{{\rm out},t},3),\nonumber\\
 h_t&=\beta_h h_{t-1}+(1-\beta_h)s_t,
 \quad\beta_h=0.95.
 \label{eq:history}
\end{align}
After calibration, positive observations update the history; zero outputs leave it unchanged.
Each observation uses the same scale as the current transform.
The updated history sets the next threshold.

\subsection{Implementation}

The first 40 successful optimizer updates use global clipping at norm 1 to initialize moments and histories.
A selected matrix enters the spectral branch when its previous second moment, history, and participation count are ready.
Other tensors use a threshold $\lambda f$, where $f$ tracks their clipped output norms.
These tensors include the MoE experts and embeddings, so the output-norm controller covers a large part of the model.
Appendix~\ref{app:state} specifies initialization, intermittent participation, and checkpoint restoration.

The distributed implementation all-gathers gradient fragments into complete logical matrices, applies the transformation, and writes each rank's local optimizer slice.
Stored row and column summaries recover the previous second-moment scale without repeatedly gathering the full moment tensor.
The transform uses FP32 within BF16 training\citep{micikevicius2017mixed}.
For $q=\min(m,n)$, its matrix arithmetic costs $O(\max(m,n)q^2+Kq^3+pmn)$, with $K=5$ and $p=3$.
Section~\ref{sec:efficiency} reports the time per step observed during training.

\begin{table*}[!t]
\centering
\begin{tabular}{lrrr}
\toprule
Method & Train loss $\downarrow$ & Validation loss $\downarrow$ & CAST lower \\
\midrule
\FullMainRows
\bottomrule
\end{tabular}
\caption{Pretraining at 5,000 steps (83.89B nominal tokens). Train: mean over steps 4801--5000. Validation: unweighted mean over 20 subsets at step 5000. The final column counts subsets where CAST has lower loss. All runs use seed 48.}
\label{tab:main}
\end{table*}

\begin{figure*}[!t]
\centering
\includegraphics[width=\textwidth]{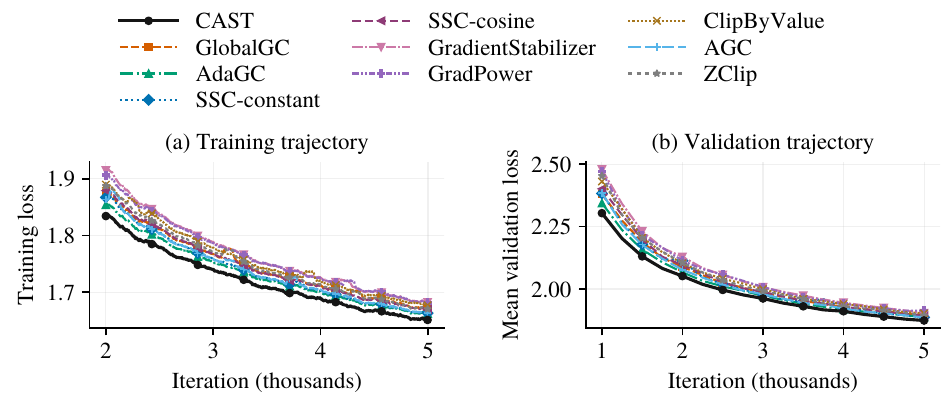}
\caption{Quality trajectories for all ten primary conditions. Training curves show trailing 100-step means from iteration 2000 onward; validation curves average the same 20 subsets at each complete checkpoint. Appendix~\ref{app:additional} includes the training trajectory from initialization.}
\label{fig:quality}
\end{figure*}

\section{Experiments}
\label{sec:experiments}

We evaluate pretraining quality, component effects, and training cost over 5,000 steps (83.89B nominal tokens).

\subsection{Experimental Setup}
\label{sec:setup}

\paragraph{Model and recipe.}
We train a shortcut MoE model\citep{cai2024shortcut} with 10 blocks, hidden size 1536, and 16 attention heads\citep{vaswani2017attention}.
Each block contains two multi-head latent attention modules\citep{liu2024deepseekmla}, two dense SwiGLU modules, and one MoE module\citep{shazeer2017outrageously,fedus2022switch,jiang2024mixtral}.
The model structure gives approximately 19.08B total parameters.
Its target activated backbone has 647.6M parameters, excluding input and output embeddings.
Actual expert activation depends on routing.
The full calculation appears in Appendix~\ref{app:model}.

The runs share the recorded data and tokenizer configurations.
We use sequence length 8192, global batch size 2048, microbatch size 1, and training seed 48.
AdamW uses $(\beta_1,\beta_2)=(0.9,0.95)$, $\epsilon=10^{-16}$, and weight decay 0.1.
The learning rate reaches $8\times10^{-4}$ after 500 warmup iterations and remains constant through the comparison window.
Training uses BF16 and expert parallelism of eight.
The model, optimizer fields, and recorded learning-rate trajectory agree across the conditions.

\paragraph{Baselines.}
We compare with GlobalGC, AdaGC, SSC-constant, SSC-cosine, GradientStabilizer, GradPower, ClipByValue, AGC, and ZClip.
Appendix~\ref{app:baselines} gives each baseline's processing order and hyperparameters.

Both SSC baselines use five Newton--Schulz iterations on the same selected matrices in raw coordinates.
SSC-constant uses $\tau=0.1$.
SSC-cosine decays $\tau$ from 10 to 0.1 over 5,000 steps.
The component study adds three variants of CAST.

\paragraph{Measurements.}
Training loss is the mean language modeling loss over iterations 4801--5000.
We average validation loss over the same 20 subsets at step 5000.
These include domain corpora, question--answer text, and code-formatted text.
We use the recorded token-loss definition for each subset.
The 100- and 500-step training windows provide a sensitivity check.
Each condition has one run with seed 48; subsets and time points describe within-run variation.

\begin{table}[!htb]
\centering
\begin{tabular}{@{}lrrr@{}}
\toprule
Condition & \shortstack{Train\\loss} & \shortstack{Validation\\loss} & \shortstack{Gate\\fraction} \\
\midrule
\FullAblationRows
\bottomrule
\end{tabular}
\caption{CAST component study (HS: historical scaling). Loss windows follow Table~\ref{tab:main}; bold marks the lowest values. Gate fraction is the mean share of eligible matrices entering SSC over steps 4801--5000.}
\label{tab:ablation}
\end{table}

\begin{figure*}[t]
\centering
\includegraphics[width=\textwidth,trim=0bp 4bp 0bp 4bp,clip]{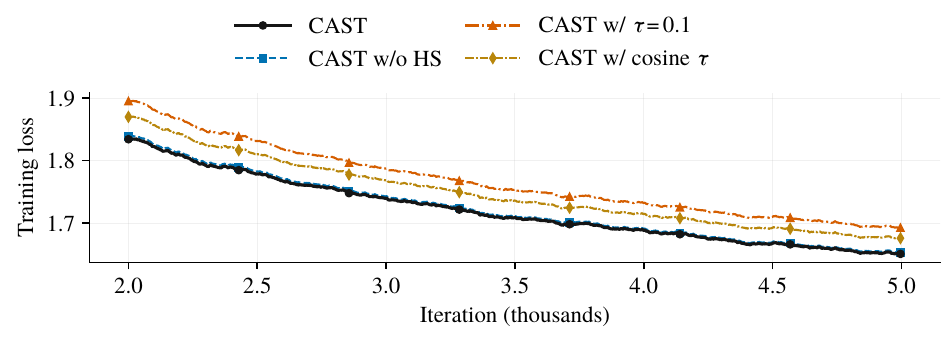}
\caption{Training trajectories for CAST and its three component variants (HS: historical scaling). Curves show trailing 100-step means from iteration 2000 onward; Table~\ref{tab:ablation} reports their final training and validation losses.}
\label{fig:ablation}
\end{figure*}

\subsection{Pretraining Results}

CAST achieved the lowest training and mean validation losses among the nine baselines at step 5000 (Table~\ref{tab:main}).
CAST's mean validation loss was $1.26\times10^{-2}$ below SSC-constant and $2.10\times10^{-2}$ below SSC-cosine.
CAST had lower loss on 20 and 19 subsets, respectively.
SSC-cosine had higher training and mean validation losses than SSC-constant.
AGC had the closest mean validation loss, with a gap of $1.18\times10^{-2}$.
CAST's training loss remained lowest across the 100-, 200-, and 500-step tail windows.

Figure~\ref{fig:quality} shows the loss curves.
Appendix~\ref{app:domains} reports the losses for all 20 validation subsets.

\begin{figure*}[!t]
\centering
\includegraphics[width=\textwidth,trim=0bp 0bp 0bp 16bp,clip]{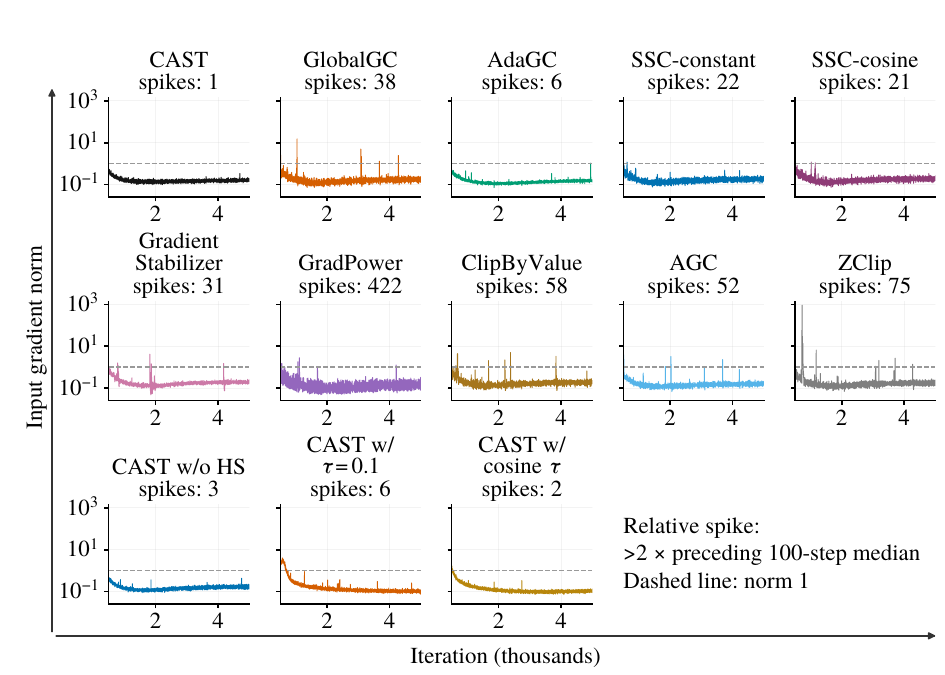}
\caption{Raw input-gradient norms over iterations 501--5000 for ten primary conditions and three CAST ablations. The final row removes HS (historical scaling) or replaces the spectral EMA with fixed or cosine thresholds. All panels share a logarithmic vertical scale. Titles report relative spike counts; the dashed line marks norm 1. Table~\ref{tab:stability} gives the statistics.}
\label{fig:gradient}
\label{fig:allgrad}
\end{figure*}

\begin{table*}[!t]
\centering
\begin{tabular}{lrrrrr}
\toprule
Method & \shortstack{Median\\norm} & \shortstack{P99\\norm} & \shortstack{Maximum\\norm} & \shortstack{Steps with\\norm $>1$} & \shortstack{Relative\\spikes} \\
\midrule
\FullStabilityRows
\bottomrule
\end{tabular}
\caption{Input-gradient norms before clipping, steps 501--5000. P99: 99th percentile. Relative spikes exceed twice the preceding 100-step median, counting consecutive steps. Bold/underline: lowest/second-lowest counts, including ties. Last three rows: CAST ablations (HS: historical scaling).}
\label{tab:stability}
\end{table*}

\subsection{Ablation Study}
\label{sec:ablation}

Table~\ref{tab:ablation} compares CAST with three component variants.
The variant without HS sets $S=\mathbf{1}$, retaining the spectral EMA and AdamW's moments.
The other variants keep $S$ and use fixed $\tau=0.1$ or cosine $\tau$ in standardized coordinates.
All three keep CAST's calibration and output-norm controller for unselected tensors.

The variant without HS increased training and mean validation losses by $2.42\times10^{-3}$ and $8.96\times10^{-4}$, respectively.
CAST had lower loss on 18 validation subsets.

Replacing the fixed threshold with cosine decay lowered training and mean validation losses by $1.70\times10^{-2}$ and $1.82\times10^{-2}$.
CAST's mean validation loss was $2.96\times10^{-2}$ lower than the cosine-threshold variant's.
Both threshold variants activated SSC for every eligible matrix in the tail window.

\begin{table}[!b]
\centering
\setlength{\tabcolsep}{4pt}
\begin{tabular}{@{}lccc@{}}
\toprule
Measure & CAST & SSC-constant & ClipByValue \\
\midrule
\FullEfficiencyRows
\bottomrule
\end{tabular}
\caption{Iteration times over steps 4801--5000 (200 steps per method, 64 devices). Mean is the arithmetic average; Median and P95 are the 50th and 95th percentiles. Throughput is $2048\times8192/(64\bar t)$, where $\bar t$ is the mean iteration time in seconds. Timings include execution and logging.}
\label{tab:efficiency}
\end{table}

\subsection{Training Stability}

Input-gradient spikes mark abrupt increases before clipping.
Figure~\ref{fig:gradient} shows these norms after learning-rate warmup.
We count norms above 1 and relative spikes.
A relative spike exceeds twice the preceding 100-step median, including pre-window points when needed.
Both counts include consecutive exceedances.
Relative spikes track sudden increases against each run's recent gradient scale.
The absolute count applies the same norm threshold across runs.

CAST recorded one relative spike and no input norm above 1 during iterations 501--5000.
SSC-constant and SSC-cosine recorded 22 and 21 relative spikes, respectively (Table~\ref{tab:stability}).
CAST and SSC-constant had similar median norms, while CAST had a smaller upper tail.

Among the baselines, GradPower had the lowest median norm but 422 relative spikes.
ZClip had the largest isolated norm, at 918.79, and 75 relative spikes.
AdaGC had six relative spikes and a lower 99th-percentile norm than CAST.
CAST combines the fewest relative spikes with the lowest mean validation loss.

Cosine decay reduced the fixed-threshold variant's relative spikes from six to two and norms above 1 from 194 to 21.

\subsection{Training Efficiency}
\label{sec:efficiency}

\begin{figure}[!ht]
\centering
\includegraphics[width=\columnwidth]{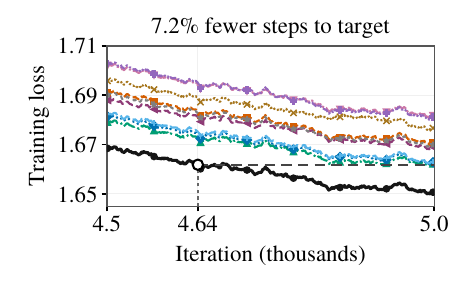}
\caption{Final 500 steps for CAST and nine baselines (100-step means). CAST reaches AdaGC's final loss (dashed) at step 4,639. Colors follow Figure~\ref{fig:fulltraining}.}
\label{fig:efficiency_zoom}
\end{figure}

Table~\ref{tab:efficiency} compares 64-device runs on the same hardware type.
Each device has about 60\% of an NVIDIA H100's compute capacity.
CAST takes 2.0\% longer per step than SSC-constant and 1.9\% longer than ClipByValue.
Timings include execution and logging and vary with cluster load.
AdaGC has the lowest final smoothed loss among the baselines.
CAST first reaches this loss at step 4,639, using 7.2\% fewer steps (Figure~\ref{fig:efficiency_zoom}).
At the mean step times in Table~\ref{tab:efficiency}, reaching AdaGC's loss target takes an estimated 5.5\% less time than 5,000 ClipByValue steps.

\section{Conclusion}

\method{} uses two histories to guide spectral clipping in pretraining.
AdamW's previous second moment sets the gradient scale.
An EMA of the clipped output's estimated spectral norm sets each matrix's threshold.
In 5,000-step MoE pretraining, CAST achieved lower training and mean validation losses than nine baselines and three component variants.
It also had the fewest relative gradient spikes.
CAST reached AdaGC's final smoothed loss in 7.2\% fewer steps.
Including per-step overhead, the estimated time saving is 5.5\% relative to 5,000 ClipByValue steps.
These results support using both histories to improve pretraining quality and efficiency.
\label{sec:main-end}

\section*{Limitations}

Full-matrix computation and communication times depend on tensor shape.
The reported timings also reflect the cluster environment.
Changes in hardware or training settings can also change the exact speedup.
The timing comparison uses the same model, training settings, hardware type, and device count.
This shared setup supports a fair comparison, so the estimated relative speedup remains a useful reference.

\nocite{liu2025muon}
\nocite{robbins1951stochastic}
\bibliography{custom}

\appendix
\section{Numerical Details}
\label{app:numerics}

\subsection{Orientation and Normalization}

For a standardized gradient $Z\in\mathbb{R}^{m\times n}$, the implementation forms a full matrix $B$ with $q=\min(m,n)$ rows.
It uses $B=Z$ when $m<n$, and $B=Z^\top$ otherwise, including square matrices.
The output is transposed back before restoring $S$.
This convention makes the left Gram $BB^\top$ the smaller Gram.

The Newton--Schulz routine scales its inputs before forming the Gram matrix.
Let
\begin{align}
 d_*&=\max\{\tau,\max_{i,j}|B_{ij}|\},\nonumber\\
 B_0&=B/d_*,\qquad t_0=\tau/d_*,\nonumber\\
 C_0&=B_0B_0^\top,\qquad
 u_0=\min\{\|C_0\|_{\rm F},\|C_0\|_\infty\},
\end{align}
where $\|C_0\|_\infty$ is the maximum absolute row sum.
The gate returns an unchanged gradient copy when $u_0\leq t_0^2$.
In the active branch, define
\begin{equation}
 \alpha=t_0^2+u_0,\qquad A_0=(t_0^2I+C_0)/\alpha.
\end{equation}
Starting with $(Y_0,R_0)=(A_0,I)$, the coupled iteration is
\begin{align}
 T_j&=\tfrac12(3I-R_jY_j),\nonumber\\
 Y_{j+1}&=Y_jT_j,\qquad R_{j+1}=T_jR_j.
 \label{eq:coupled}
\end{align}
The transformed matrix is
\begin{equation}
 \widetilde B=(t_0/\sqrt\alpha)R_5B.
 \label{eq:nsoutput}
\end{equation}
The rightmost factor is the original full matrix $B$.
An unrepresentable positive normalized threshold or a nonfinite intermediate raises a numerical error before state commit.

\subsection{Spectral Norm Estimation}

For a nonzero output $X$, $\PowerTop(X,3)$ starts from a normalized Gaussian vector $x$.
Each round normalizes $Xx$ and then $X^\top(Xx/\|Xx\|_2)$.
After three rounds, the observation is $\|Xx\|_2$.
The vector is seeded by the logical parameter identity, successful update index, and probe purpose, using base seed 1234.
A degenerate normalization triggers one retry with a different seeded probe.
A zero matrix supplies no positive observation.

\section{State and Communication}
\label{app:state}

Table~\ref{tab:controller} lists the controller settings.

\subsection{Parameter Selection}
\label{app:scope}

Selection uses each complete logical parameter before optimizer sharding is restored.
The spectral branch selects two-dimensional logical parameters in the dense optimizer group.
It excludes parameters whose lowercase names contain an embedding/output exclusion marker.
Those markers include \texttt{embedding}, \texttt{lm\_head}, and \texttt{output\_layer}; position embeddings are also excluded.
The expert-matrix and embedding-matrix inclusion switches are both false.
MoE projections and routers follow the same selection rule.
Routing tensors assigned to the expert optimizer follow the norm branch even when their logical shape is two-dimensional.
All other shapes also use the norm branch.

\subsection{Calibration and State Updates}

The first 40 successful global updates apply GlobalGC with norm cap 1.
Each logical tensor separately counts its successful participating updates.
Spectral observations begin when a valid previous second moment is available.
The first positive observation initializes $h$ directly.
Subsequent calibration observations use the ordinary EMA.
The spectral branch becomes active once its history and previous second moment are ready and calibration is complete.
A late-participating tensor uses norm cap 1 until ready.

For a tensor $T$ outside the spectral branch, the active transformation is
\begin{equation}
 T_{\rm out}=T\min\{1,\lambda f/\|T\|_{\rm F}\},
 \label{eq:normfallback}
\end{equation}
with zero output for zero input.
Positive output norms update $f$ with the same initialization and decay rules as $h$.
The histories have floor $10^{-12}$.
There is no additional global cap after the active branches.

If an active tensor returns after a gap of $\Delta$ successful global updates, its history uses decay $\beta_h^\Delta$.
A missing gradient leaves state unchanged.
A genuine zero gradient can advance participation without providing a positive history observation.
Moment update counts, participation counts, and observation indices remain distinct.
The optimizer stages transformed gradients and candidate controller states, then calls AdamW once.
It commits the controller states only when that update succeeds.
Checkpoints preserve these histories, counts, factored moment summaries, and a configuration fingerprint.

Gradient fragments are packed into buckets and all-gathered within the relevant data-parallel or expert-data-parallel group.
Each rank reconstructs and transforms the same complete logical matrix, then writes its local optimizer slice.
Full-matrix temporaries and communication therefore remain part of the implementation cost.
The norm branch obtains its magnitudes through scalar reductions.

The row summary of the uncorrected moment obeys
\begin{equation}
 \bar r_{i,t}=\beta_2\bar r_{i,t-1}
 +(1-\beta_2)\frac1n\sum_j(G_{{\rm out},t})_{ij}^2.
\end{equation}
Column and global summaries follow the same update rule.
Linearity makes these the corresponding means of AdamW's second moment in exact arithmetic.
Bias correction recovers the scale inputs in Equation~\ref{eq:scale}.
The optimizer still stores and uses its full moments.
The summaries support clipping and avoid repeatedly gathering the full previous second moment.

\begin{table}[t]
\centering
\begin{tabular}{ll}
\toprule
Controller setting & Value \\
\midrule
Selected spectral space & Full matrix \\
Newton--Schulz iterations & 5 \\
Output power iterations & 3 \\
Multiplier $\lambda$ & 1.3 \\
History decay $\beta_h$ & 0.95 \\
Calibration updates / cap & 40 / 1 \\
Scale floor $\rho$ & 0.01 \\
Scale epsilon $\epsilon_s$ & $10^{-8}$ \\
History floors & $10^{-12}$ \\
Probe seed & 1234 \\
Additional global clipping & None \\
\bottomrule
\end{tabular}
\caption{Main-method controller settings. The training seed is separately fixed at 48.}
\label{tab:controller}
\end{table}

\section{Model and Baselines}
\label{app:model}

\subsection{Parameter Counts}

\begin{table*}[!t]
\centering
\begin{tabular}{lr}
\toprule
Parameter component & Count across the model \\
\midrule
20 MLA modules, including low-rank normalization & 114,698,240 \\
20 dense SwiGLU modules & 283,115,520 \\
Learned experts & 18,119,393,280 \\
MoE input/output projections & 47,185,920 \\
Routers & 13,762,560 \\
Block and MoE normalization & 76,800 \\
Input embedding and output projection & 503,316,480 \\
Final normalization & 1,536 \\
\midrule
Total & 19,081,550,336 \\
Target activated backbone, excluding embedding/output & 647,584,256 \\
\bottomrule
\end{tabular}
\caption{Parameter counts from the recorded architecture and model source. Replicated parameters are counted once. The activated backbone uses the target of 32 learned experts per MoE module; the actual count varies across tokens. Selecting 48 learned experts per MoE module gives an upper bound of approximately 0.742B activated backbone parameters.}
\label{tab:model}
\end{table*}

\begin{figure*}[!t]
\centering
\includegraphics[width=\textwidth]{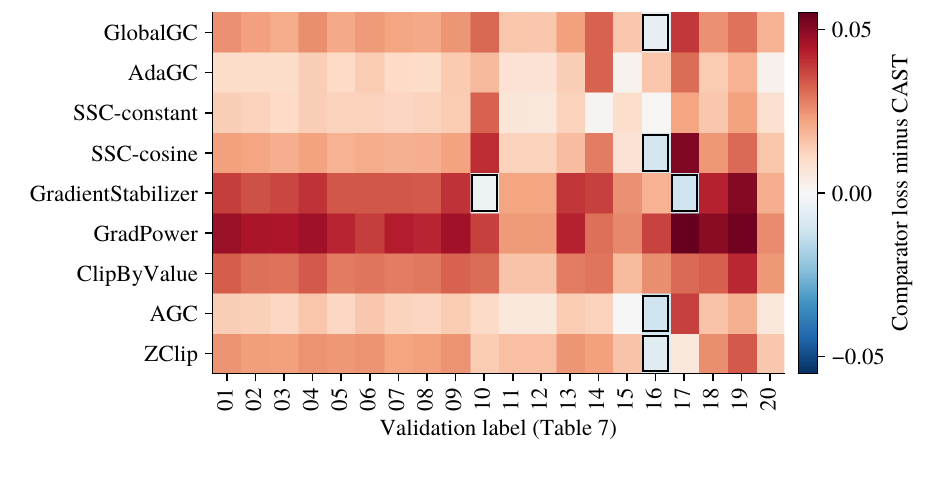}
\caption{Every baseline-minus-\method{} validation loss difference. Positive values favor \method{}; outlined negative cells favor the baseline. Subsets correspond to Table~\ref{tab:domains}.}
\label{fig:domains}
\end{figure*}

\begin{figure*}[t]
\centering
\includegraphics[width=\textwidth]{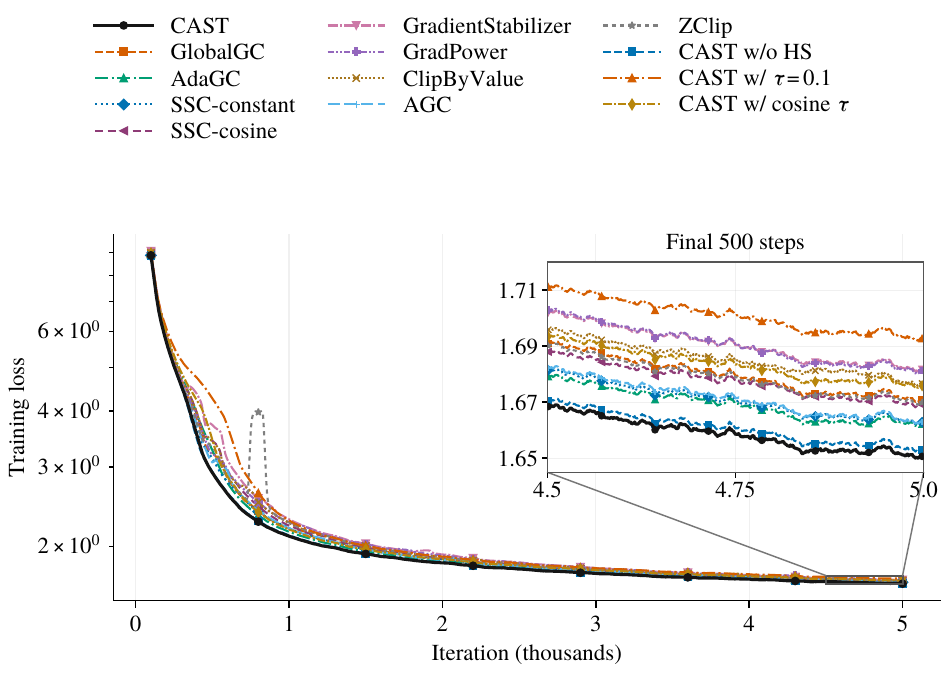}
\caption{Training from the first complete 100-step averaging window through iteration 5000. All ten primary conditions and three CAST ablations (HS: historical scaling) use trailing 100-step means. The main plot uses a logarithmic vertical axis; the box and connected inset show the final 500 steps on a linear scale.}
\label{fig:fulltraining}
\end{figure*}

Each of the 10 shortcut blocks contains two MLA attention modules, two dense SwiGLU modules, and one MoE module.
The hidden width is 1536 and the dense feed-forward width is 3072.
Each MLA has 16 heads, query low-rank width 384, and key/value low-rank width 128.
Its non-RoPE query/key width is 128 per head, RoPE width is 64, and value width is 128.
These dimensions give 5,734,912 parameters per MLA, including its low-rank normalization parameters.

Each MoE module contains 3,072 learned experts and 512 identity experts without feed-forward weights.
Learned experts operate on 384-dimensional inputs with feed-forward width 512.
Routing selects 48 candidates in total and targets 32 learned-expert activations per token; the actual learned-expert count varies with routing.
The input embedding and output projection are untied, with vocabulary size 163840.
Table~\ref{tab:model} summarizes the parameter counts.

\subsection{Baseline Settings}
\label{app:baselines}

All transformations precede both AdamW moment updates.
GlobalGC uses a global norm cap of 1.
AdaGC uses per-tensor multiplier 1.04, EMA decay 0.99, and 100 calibration steps.
These values come from AdaGC's run configuration.

SSC-constant applies Equation~\ref{eq:ssc} to the selected dense matrices in raw coordinates, with threshold 0.1 and five Newton--Schulz iterations.
It starts immediately and passes all unselected tensors through unchanged.
For square gradients, it forms the Gram matrix from the untransposed input.
SSC-cosine replaces the fixed threshold with
\begin{equation}
\begin{aligned}
 \tau_k &= 0.1+4.95\bigl[1+\cos(\pi u_k)\bigr],\\
 u_k &= \min\{(k-1)/4999,1\}.
\end{aligned}
\label{eq:cosine}
\end{equation}
Here $k$ is the one-based training iteration, including skipped updates.
The threshold starts at 10 and reaches 0.1 at iteration 5000.
It stays at 0.1 afterward.

GradientStabilizer uses global norm and squared-norm EMA decays 0.6 and 0.999, without bias correction or an extra global cap.
GradPower applies GlobalGC(1), then the signed elementwise power with exponent 1.2.
ClipByValue clamps each coordinate to $[-0.1,0.1]$ without another clipping step.
AGC uses multiplier 0.01, weight-norm floor $10^{-3}$, and gradient-norm floor $10^{-6}$, including embedding and output tensors.
ZClip uses decay 0.97, z-score threshold 2.5, 25 calibration steps, and epsilon $10^{-6}$.
It applies an additional norm cap of 1 without skipping spike steps.

The three component variants preserve CAST's calibration and output-norm controller for unselected tensors.
The variant without HS removes the clipping scale $S$ while retaining AdamW's second moment.
The fixed-threshold variant sets the selected spectral threshold to 0.1 while retaining HS.
The cosine-threshold variant uses Equation~\ref{eq:cosine} in the same standardized coordinates.
Its schedule follows the training iteration throughout calibration and checkpoint resumes.

\section{Validation Results}
\label{app:domains}

Table~\ref{tab:domains} includes every baseline and validation label.
Question--answer and benchmark-formatted rows measure the configured token loss for those texts.
Figure~\ref{fig:domains} shows the loss differences for each baseline and validation subset.

\begin{table*}[p]
\centering
\begin{tabular}{lrrrrr}
\toprule
Label & \method{} & GlobalGC & AdaGC & SSC-constant & SSC-cosine \\
\midrule
\FullDomainRowsA
\bottomrule
\end{tabular}

\medskip
\begin{tabular}{lrrrrr}
\toprule
Label & GS & GradPower & Value & AGC & ZClip \\
\midrule
\FullDomainRowsB
\bottomrule
\end{tabular}
\caption{Token LM losses on all 20 validation subsets at iteration 5000. Bold and underline mark the best and second-best values across all ten conditions, spanning both panels. GS is GradientStabilizer; Value is ClipByValue.}
\label{tab:domains}
\end{table*}

\section{Full Training Curves}
\label{app:additional}

The full training overview is shown in Figure~\ref{fig:fulltraining}.
The main-text Figure~\ref{fig:gradient} and Table~\ref{tab:stability} contain the input-gradient traces and statistics for these conditions.

\end{document}